\documentclass[final,journal,letterpaper]{IEEEtran}

 \usepackage[hidelinks]{hyperref}
 \usepackage{varioref} %smart page, figure, table, and equation referencing
 \usepackage{wrapfig} %wrap figures/tables in text (i.e., Di Vinci style)
 \usepackage{threeparttable} %tables with footnotes
 \usepackage{dcolumn} %decimal-aligned tabular math columns
  \newcolumntype{d}{D{.}{.}{-1}}
 \usepackage{nomencl} %nomenclature generation via makeindex
  \makeglossary
 \usepackage{subfigmat} % matrices of similar subfigures, aka small mulitples
 \usepackage{fancyvrb} %extended verbatim environments
  \fvset{fontsize=\footnotesize,xleftmargin=2em}
 \usepackage{lettrine} %dropped capital letter at beginning of paragraph
 \usepackage{algorithm,algorithmic}

 \usepackage{epsfig}
 \usepackage{epstopdf}
 \usepackage{subfigure}
 \usepackage{latexsym}
 \usepackage{amsmath}
 \usepackage{amsfonts}
 \usepackage{amssymb}
 \usepackage{mathrsfs}
 \usepackage{cases}
 \usepackage{array}
 \usepackage{color}
 \usepackage{soul}
 \setulcolor{red}
 \sethlcolor{yellow}
 \newcommand{\comment}[1]{}
 \usepackage{url}
 \usepackage[noadjust]{cite}
 \usepackage{graphicx}
 \usepackage{psfrag}
 \usepackage{color}
   \usepackage{multirow}
 \usepackage{ragged2e}
 \usepackage{graphicx}
 \usepackage{RL_notation}

\begin{document}
\title{Learning and Fast Adaptation for Grid Emergency Control via Deep Meta Reinforcement Learning}

\author{Renke Huang,~\IEEEmembership{Senior Member,~IEEE}, Yujiao Chen,~\IEEEmembership{Member,~IEEE}, Tianzhixi Yin,
       Qiuhua Huang,~\IEEEmembership{Member,~IEEE,}
        Jie Tan, Wenhao Yu, Xinya Li, Ang Li, Yan Du,~\IEEEmembership{Member,~IEEE,}% <-this % stops a space
        
\thanks{The Pacific Northwest National Laboratory is operated by Battelle for the U.S. Department of Energy (DOE) under Contract DE-AC05-76RL01830. This work was supported by DOE ARPA-E OPEN 2018 Program. \newline
$\ast$ \textit{Corresponding author: Qiuhua Huang}}

\thanks{R. Huang, Q. Huang, T. Yin, X. Li, A. Li, Y. Du are with Pacific Northwest National Laboratory, Richland, WA 99354, USA (e-mail: \{renke.huang, qiuhua.huang, tianzhixi.yin, xinya.li, ang.li, yan.du\}@pnnl.gov).}
\thanks{Yujiao Chen was a post-doc with PNNL.}
\thanks{Jie Tan, Wenhao Yu are with Google Research, Google Inc, Mountain View, CA, 94043 USA (e-mail: \{jietan, magicmelon\}@google.com).}}

\maketitle

\begin{abstract}
As power systems are undergoing a significant transformation with more uncertainties, less inertia and closer to operation limits, there is increasing risk of large outages. Thus, there is an imperative need to enhance grid emergency control to maintain system reliability and security. Towards this end, great progress has been made in developing deep reinforcement learning (DRL) based grid control solutions in recent years. However, existing DRL-based solutions have two main limitations: 1) they cannot handle well with a wide range of grid operation conditions, system parameters, and contingencies; 2) they generally lack the ability to fast adapt to new grid operation conditions, system parameters, and contingencies, limiting their applicability for real-world applications. In this paper, we mitigate these limitations by developing a novel deep meta-reinforcement learning (DMRL) algorithm. The DMRL combines the meta strategy optimization together with DRL, and trains policies modulated by a latent space that can quickly adapt to new scenarios. We test the developed DMRL algorithm on the IEEE 300-bus system. We demonstrate fast adaptation of the meta-trained DRL polices with latent variables to new operating conditions and scenarios using the proposed method, which achieves superior performance compared to the state-of-the-art DRL and model predictive control (MPC) methods.  

\end{abstract}


% \fi
\IEEEpeerreviewmaketitle

\section{Introduction}


\IEEEPARstart {P}{OWER} systems are facing increased risks of large outages due to main factors including aging infrastructure, significant change of generation/load mix \cite{AEMO2017report}, extreme weather\cite{kenward2014blackout}, threats from physical and cyber attacks. This is evident by increasing occurrence of large outages \cite{kenward2014blackout,Bo2015}. In this context, corrective and emergency control is imperative in real-time operation to minimize the occurrence and impact of power outages and blackouts \cite{preventive_emergency_control, Huang2020_DRL}. Furthermore, improved emergency control capabilities help the industry reduce overly relying on, thus the cost of, preventive security measures, thereby achieving overall better efficiency and asset utilization \cite{strbacemerging}.

%Conventional ruled-based control methods such as under-voltage load shedding protections are not adaptive and tend to be over-conservative\cite{van2007vs_overview}, while optimization-based methods such as model-predictive control (MPC) are computationally expensive and thus not suitable for solving large power system control problems, particularly for real-time control\cite{Glavic2006_MPC_voltage}. 

Recently, deep reinforcement learning (DRL) made outstanding progress \cite{Silver2017,li2017_DRL_review} and showed promising results for providing fast and effective power system stability and emergency control solutions \cite{glavic2019_DRL4GridReview,Huang2020_DRL,cao2020reinforcement}. The standard formulation of reinforcement learning (RL) is to maximize the average (or expected) accumulative rewards over the considered scenarios of the environments \cite{sutton2018_RLBook}. However, for power systems with many significantly different operation conditions and increased uncertainties, the policy learned by DRL may not work well when the grid operation condition changes notably at the testing or deployment stage, leading to unsatisfactory or even unacceptable system performance and outcomes. 

%One approach for addressing this issue is to train multiple DRL agents and each is responsible for a subset of potential (or forecast) operation conditions and/or fault scenarios. However, this will cost more training time and require more computing resources. In addition, the RL agents tend to have limited generalization capability due to fewer scenarios that they are exposed to during training. 
One approach to partially addressing the issue is moving the RL training as close as possible to the real operation time to reduce the uncertainties (thus the variance of operation conditions has to be considered) by shortening the required training  time. We developed the parallel augmented random search (PARS) algorithm  that is highly scalable to accelerate the training in our previous work\cite{huang2020accelerated_DRL}. However, a  fundamental yet generally missing capability for RL agents (controllers) is to quickly adapt to new grid operation conditions. Indeed, researchers in \cite{Wehenkel2020_ML4RM} pointed out ensuring the adapability of machine learning models is a requirement for practical acceptance of machine learning applications. The adaptability problem in the power system emergency control area has not been addressed yet. In this paper, we tackle this problem by developing a novel deep meta-reinforcement learning (DMRL) algorithm and applying it to learn and adapt power system emergency control policies against voltage stability issues. The proposed DMRL algorithm can quickly adapt the behavior of the trained policies to unseen scenarios in a new target environment. It combines the meta strategy optimization together with DRL, and trains a policy modulated by a latent space that can quickly adapt to new and much different grid operation conditions, and system parameters. Given fast-changing operation conditions and increased uncertainties in power systems, we believe this work is a major advancement for DRL application for grid control. 
\vspace{-10pt}
\subsection{Literature review}

There are a number of recent efforts in applying RL (particularly DRL) for power system stability and emergency control \cite{yan2018data,yan2020multi,li2020hierarchical,diao2019autonomous,glavic2019_DRL4GridReview,chen2020model, Huang2020_DRL}. However, there are some notable limitations among them: (1) they trained a single policy to handle many different power grid environments with the so-called domain randomization technique \cite{tobin2017domain} and simply relied on the unwarranted generalization capability of deep neural networks to generalize to unseen grid environments, thus the trained polices may not work well if the grid environment changes drastically; (2) they lacked the capability of fast adaptation to new grid environments by leveraging prior learning experience.

In machine learning literature, existing works in adapting control policies to new environments are mainly in the following two categories: 1) model-based adaption methods; 2) model-free adaptation methods.  In model-based adaptation methods\cite{nagabandi2018learning,lenz2015deepmpc}, a dynamic model is first learnt using some data-driven techniques to capture or represent the system dynamics in the training environment, and then the learnt dynamic model is adapted with some recent observations from the target environment, and finally the control policy is determined using methods such as model predictive control (MPC)\cite{lenz2015deepmpc}. These methods can adapt to changes in the environment on-line. However, there are two main shortcomings in terms of application for large-scale power system emergency control: 1) learning good dynamic models of large-scale, highly non-linear power systems is highly challenging, and has yet proven practically feasible; 2) solving large-scale MPC to obtain optimal control policy is computationally expensive\cite{huang2020accelerated_DRL}, and thus almost impossible to meet the real-time requirement of power grid emergency control.    

In model-free adaption methods, the control policy (modeled as a neural network) is directly adjusted according to the observed dynamics from the targeted environment. One popular class of such methods is the gradient-based meta learning approach \cite{finn2017maml,finn2017meta}. However, gradient-based approaches usually require small incremental parameter adjustment to stabilize the learning process or ensure numerical stability \cite{botvinick2019_RL_fast_slow}.  Thus, without a large amount of experience (adaption steps) to update the control policy parameters, the meta-learnt control policy could not be significantly adjusted to adapt to a quite different environment. 

Another class of the model-free methods is latent space-based adaptation method \cite{rakelly2019efficient, yu2020learning}. It encodes the training experience into a latent context (space), and the control policy is conditioned on the latent context. The latent context is then fine-tuned for new environments. Most efforts in this line of research are to first train an inference model in conjunction with the RL training process during the training stage, and during adaption, leverage the inference model to infer the latent context through the observation input from the target environment. However, when the target or actual environment differs notably from those considered in the training, the inference model may produce non-optimal results, leading to poor adaptation performance. As pointed out in \cite{yu2020learning}, this is mainly due to the fact that the process of learning latent context in training and inferring the suitable latent context in adaptation are not consistent.  By extending the same latent space optimization process from the meta-learning stage to both meta-learning and adaptation stages to overcome such discrepancy, Yu {\em{et~al.}\ }\cite{yu2020learning} developed the so-called meta strategy optimization (MSO) method and showed that it allows the agents to learn better latent space that is suitable for fast adaptation to new environments. In our work, we adapted this technique for power system emergency control problems. 
\vspace{-6pt}
\subsection{Contributions}

Our contributions in this paper include the following:

\begin{enumerate}

    \item We develop a novel DMRL algorithm that combines the PARS \cite{huang2020accelerated_DRL} and MSO \cite{yu2020learning} algorithms, which could train an agent that can quickly and effectively adapt to a new environment. 
    \item The developed DMRL algorithm consists of a meta-learning stage and an adaptation stage (see Fig. \ref{fig:metalearning4grid}), which naturally fits into the execution time frame of existing power system operation procedures. 
    \item We apply the developed DMRL algorithm for training  power system emergency control policies against voltage stability problems and more importantly fast adapting them to unseen operation conditions and/or contingency scenarios.
    \item We demonstrate fast and effective adaptation of the meta-trained DRL polices to new operation scenarios including unseen power flow cases, new system dynamic parameters and new contingencies using the proposed method and the testing results show superior performance compared to other baseline methods.  
\end{enumerate}

While we focus on grid emergency control applications in our test cases in this paper, the proposed method is generic and could be  extended to other control and decision-making problems in power systems without much difficulty. 
\vspace{-6pt}

\subsection{Organization of the paper}

The rest of the paper is organized as follows: Section \ref{sec:prob_formul} introduces the problem formulation and our proposed method at a high level. Section \ref{sec:dmrl} presents the details of our proposed DMRL algorithm. Test cases and results are shown in Section \ref{sec:results}.
Conclusions and future work are provided in Section \ref{sec:conclusions}.

\section{Problem Statement and Proposed Method}\label{sec:prob_formul}
\comment{
{\color{red} why we need meta reinforcement learning?? we should try to define the problem here, preferably in a mathematical way, showing average over a number of scenarios\\
Use the uncertainty plot to show the problem\\
!! Compare with the alternative of a family of control policies, with each for one or few scenarios (did not leverage the underlining commonality/similarity, does not generalize well\\
Use the adaptation+uncertainty illustration plot to show the key idea of our approach\\
6

1) The nature of  RL that it optimizes the average/expected performance over a number of scenarios and/or conditions means that it is short of optimality for specific grid operation conditions, particularly when the scenarios considered in the training stage are significantly different.Examples of grid scenarios that would  benefit from fast adaptation of RL-based control: 1) an hurricane or storm is coming; 2) risk of widefire at certain regions of the grid; 3) high voltage stability risks at some load centers. In the rest of the paper, we will focus on voltage stability problems. \\

2) Comparison with training RL specifically for a few scenarios, DMRL provides better generalization capability.

}
}
We first discuss the challenges in applying DRL for grid control under fast-changing power grid operation scenarios with increased uncertainties, which necessitates and highlights the need of fast adaptation capability for DRL-based  agents or controllers. Then, we introduce meta-reinforcement learning techniques for achieving the fast adaptation. Lastly, we present the key procedures of our proposed DMRL approach and how they fit into existing power grid operation procedures.
\vspace{-6pt}
\subsection{ Challenges in Applying RL for Grid Control under Fast-Changing Operation Scenarios with Increased Uncertainties}
RL problems can be defined as policy search in a (partially observable) Markov Decision Process (MDP) defined by a tuple ($\mathcal{S,A}$,$p$,$r$,$\gamma$) \cite{sutton2018_RLBook}, where $\mathcal{S}$ is the state space, $\mathcal{A}$ is the action space, $p: \mathcal{S} \times \mathcal{A} \rightarrow \mathcal{S}$  is the transition function, and $r$: $\mathcal{S} \times \mathcal{A} \rightarrow \mathbb{R}$ is the reward function. The goal of RL is to learn a policy $\pi(s_t):\mathcal{S}\rightarrow\mathcal{A}$, such that it maximizes the expected accumulative reward $J(\pi)$ over time under $p$:
\begin{equation}
J(\pi)=\mathbb{E}_{s_0,a_0,s_1,a_1,...s_{\mathcal{T}},a_{\mathcal{T}}} \Sigma_{t=0}^{\mathcal{T}} \gamma^{t} r\left(s_{t}, a_{t}\right)
\label{eqn:RL_obj}
\end{equation}

\begin{figure}[t]
\centerline{\includegraphics[scale=0.35]{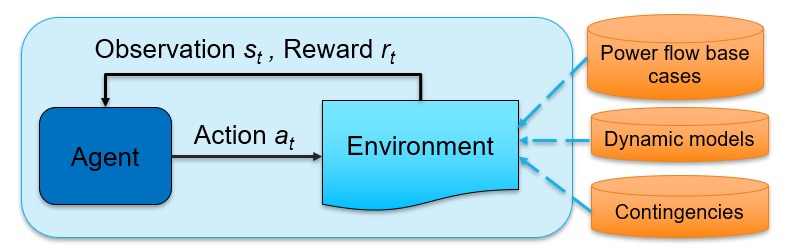}}
\caption{The agent-environment interaction in a Markov decision process}
\label{fig:RL_setup}
\vspace{-10pt}
\end{figure}

\comment{The environment transition function $\mathcal{p} :\mathcal{S} \times \mathcal{A} \rightarrow \mathcal{S}$  is the probability density of the next state $s_{t+1} \in \mathcal{S}$ given the current state $s_{t} \in \mathcal{S}$ and action $a_{t} \in \mathcal{A}$. At each interaction step, the environment returns a reward $r$: $\mathcal{S} \times \mathcal{A} \rightarrow \mathbb{R}$.  The goal of an agent is to learn a policy $\pi(s_t)$ that maximizes the objective $J(\pi)$, which is usually defined as the expected accumulative reward:

\begin{equation}
J(\pi)=\arg \max _{a_t} \mathbb{E}_{(s_t, a_t) \sim \mathcal{P}} \Sigma_{t=0}^{\mathcal{T}} \gamma^{t} r\left(s_{t}, a_{t}\right)
\label{eqn:RL_obj}
\end{equation} }
 \noindent where $a_t\sim\pi(s_t)$ and $s_{t+1}\sim p(s_t, a_t)$, and $\mathcal{T}$ is the maximum end time. Note that the accumulated reward maximization in RL is opposite to minimizing the cost objective that are usually considered in power system optimizations. A standard setup of RL problems for power grid control is shown in Fig. \ref{fig:RL_setup}. In DRL, the policy is usually parameterized by a neural network with weights $\theta$ and the policy is denote as $\pi_{\theta}$.

 %DRL is a combination of RL and deep learning technologies. The capabilities of high dimensional feature extraction and non-linear function approximation from deep learning makes it possible for DRL to directly use the raw state-space representations and train policies for complex systems and tasks in an end-to-end manner.

\begin{figure}[t]
\centerline{\includegraphics[scale=0.30]{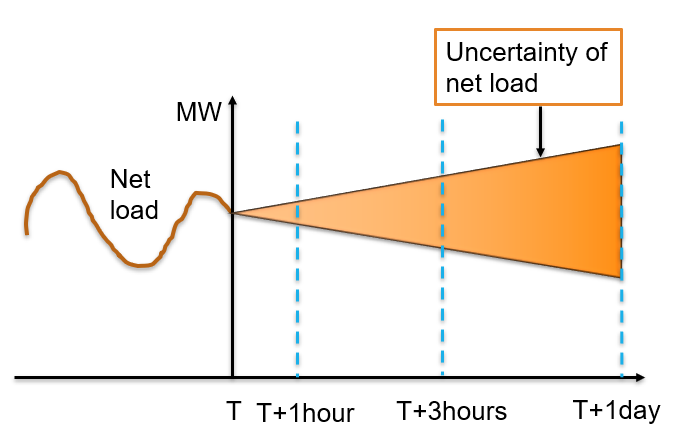}}
\caption{Uncertainties in power systems}
\label{fig:uncertainties_grid}
\vspace{-6pt}
\end{figure}

 There are increased uncertainties in power systems due to large integration of intermittent generation resources, as illustrated in Fig. \ref{fig:uncertainties_grid}. The system operation conditions could change tremendously even within a few hours (for example, the so-called duck curve \cite{hou2019probabilistic}). Since RL training  (usually ranging from hours to days) is slow compared to the required real-time emergency control time interval (i.e., within seconds), RL training has to be performed off-line, from hours to days ahead of real-time operation (denoted by time $T$ in Fig. \ref{fig:uncertainties_grid}). Since the actual operation conditions and contingency scenarios cannot be pre-determined accurately prior to their occurrence, a number of operation conditions (different power flows and system dynamic parameters) have to be considered as a distribution of different environments $\bm{P}(E_{i})$ while the potential contingencies have to be considered as a distribution of different contingencies $\bm{P}(\mathcal{C}_{j})$ when training the RL agent (or policy). The DRL algorithms typically perform well on the set of training environments $\lbrace\bm{\mathcal{E}}_{trn}\rbrace$ and they purely leverage the trained neural networks $\pi_{\theta}$ to learn representations that support generalization to a new target environment $\mathcal{E}_{tar}$. As a result, the success or effectiveness of generalization to a new environment $\mathcal{E}_{tar}$  dependent on the similarity between new target environment $\mathcal{E}_{tar}$ and the training environment set $\lbrace\bm{\mathcal{E}}_{trn}\rbrace$. In other words, the generalization may not be effective if the power grid operation conditions become quite different from those considered at the training stage.

 \comment {Otherwise, the RL model trained solely based on one single operation condition environment $E_{i}$ may not work well when tested on a different operation condition environment $E_{k}$. }

 \comment{ According to (1), a standard formulation of RL is to optimize the average/expected performance of solving a problem. Suppose the RL training is performed in a day-ahead (if not week-ahead) time frame and considers the system operation conditions of the next 24 hours, the distribution of operation condition environments $\bm{P}(E_{i})$ will most likely have a large variance (see Fig. \ref{fig:uncertainties_grid}). {\color{red}In this context, even the RL agent achieves the optimal solution of Eq (1) under these environments, it generates a single policy across the combination of different environments and scenarios $\{\bm{P}(E_{i}) \cup \bm{P}(\Gamma_{j})\}$ it sees during the training. The trained policy trades optimality for robustness: for each environment $E_i$, the solution will be sub-optimal. For some environment, the trained model may produce solutions that are too far away from the optimal solutions to be satisfactory for real-world deployment. }For example, if a few operation conditions are very difficult to solve for the RL agent, while a majority of them are easy to solve, then the trained RL model often fails to solve those difficult conditions (we show such cases in the test results). Furthermore, DRL typically specializes on the trained tasks and leverages neural networks to learn powerful representations that support generalization to new tasks. As a result, the success or effectiveness of generalization is dependent on the similarity between new task and the trained tasks. In other words, the generalization may not be effective if the power grid operation conditions become quite different from those considered in the training stage.}

\subsection{Meta-reinforcement Learning}

Meta-reinforcement learning (Meta-RL) integrates the idea of meta-learning (learning to learn) into RL. The goal of Meta-RL is to train an agent that can quickly adapt to a new task using only a few data points and training iterations. To accomplish this, the model or learner is trained at a meta-learning stage on a set of tasks, such that the trained model can quickly adapt to new tasks using only a small number of examples or trials. Formally, given a distribution of tasks $\bm{P}(T_{k})$ that we are interesting in and an adaptation process $U: \Theta \times \bm{P}(T_{k}) \mapsto \Theta$, Meta-RL finds a policy that maximizes the rewards after adaptation:
\begin{equation}
    \theta^* = \arg\max_{\theta} \mathbb{E}_{\bm{P}(T_{k})} \lbrack J(U(\theta, T_{k})) \rbrack
    \label{eqn:metarl_obj}
\end{equation}
%where $J$ is the expected accumulative reward as defined in (\ref{eqn:RL_obj}) and $\theta \in \Theta$ is the parameters of the policy.
Note that for the power grid emergency control problem we investigated in this paper, the distribution of tasks is defined by a distribution of environments presenting different system conditions and dynamic parameters, that is $\bm{P}(T_{k})=\bm{P}(E_{i})$.

An ideal Meta-RL algorithm should: 1) be data efficient during adaptation to new tasks; 2) flexibly modulate the behavior of the trained policy to achieve optimal performance for new tasks. 

\vspace{-6pt}
\subsection{Our Proposed Method}
\begin{figure}[t]
\centerline{\includegraphics[scale=0.5]{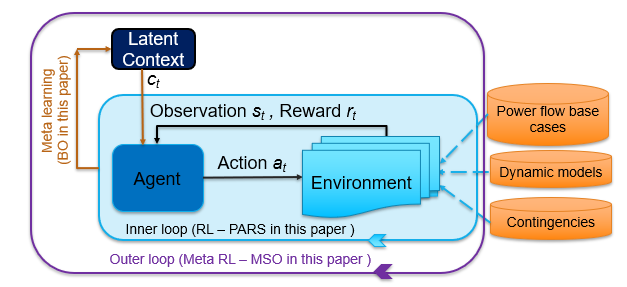}}
\caption{Latent-space based meta-reinforcement learning}
\label{fig:latent_metaRL}
\end{figure}

We propose a latent-space-based MSO algorithm to address the Meta-RL problem defined in (\ref{eqn:metarl_obj}) by extracting the training experience on the distribution of tasks $\bm{P}(T_{k}))$ into a latent space representation \cite{yu2020learning}, as illustrated in Fig. \ref{fig:latent_metaRL}.

A key idea behind MSO is that we want to obtain a \emph{universal} policy that performs optimally for every training task in $\bm{P}(T_{k})$. More importantly, this universal policy needs to be able to quickly adapt to novel tasks, which are not seen in $\bm{P}(T_{k})$. We first make the universal policy conditioned on the training tasks $\bm{P}(T_{k})$ so that the policy can learn a specialized strategy for each training task. However, training such a policy is a challenging problem because $\bm{P}(T_{k})$ is high-dimensional, which can significantly increase the input size and the complexity of the neural networks of the policy. Since this high-dimensional task space is often redundant, our method finds a lower-dimensional representation (latent space) of the training tasks $\bm{P}(T_{k})$. This learned low-dimensional representation is then used to encode the task and augment the original input of the control policy. More concretely, the proposed method defines the universal policy to be $\pi_{\theta}(s_t,c): \mathcal{S} \times \mu \mapsto \mathcal{A}$, where  $c$ is a latent vector in the low-dimensional latent space $\mu$. During training time, MSO jointly optimizes the latent space $\mu$ and the policy parameters $\theta$ such that the learned controller can effectively handle novel tasks with a small amount of fine-tuning data. In other words, MSO finds a non-linear dimension reduction of the training tasks $\bm{P}(T_{k})$, guided by the reward of the control policy \emph{after adaptation}. Given a novel task during the testing stage, our algorithm performs adaptation by identifying the task in the latent space, through finding an optimal latent vector $c$, which combines with the universal policy $\pi_{\theta}(s_t,c)$, can perform best in the test task. Since the latent space is low dimensional, the adaptation is data efficient. Furthermore, our learning algorithm trains the meta-learning model based on a set of representative forecasted system environments $\lbrace \bm{\mathcal{E}}_{trn}\rbrace$ with uncertainties being taken into account, therefore the latent representation $\mu$ can be learnt efficiently with good physical information and prior knowledge about the distributions of the system environments.

\comment {One key point to make these methods work well is that the set of training tasks needs to be properly selected to ensure that the space of policy behaviors encoded by the latent space is expressive enough to handle the desired testing scenarios. }

\comment {Existing methods in solving (\ref{eqn:metarl_obj}) can be broadly categorized into two classes that achieve different balances between the two desired properties. The first class is full-policy-based Meta-RL methods \cite{finn2017model, yang2019norml, houthooft2018evolved, rothfuss2018promp, song2020rapidly}, where the adaptation process $U$ is selected to adapt the entire set of policy parameters during testing. For instance, in Model-agnostic Meta Learning (MAML)\cite{finn2017model}, the policy parameters are adapted through a single-step policy gradient update. By modulating the full policy parameter space, these methods can induce large changes in the policy behavior, making them more flexible. However, due to the large space of parameters that need to be adapted, these methods in general require a high number of samples during adaptation to achieve optimal performance.}

\comment {\subsection{Our proposed method}

As different Meta-RL algorithms strike different trade-offs between sample efficiency and policy flexibility, choosing which algorithm to use largely depends on the requirements of the problem to be tackled. }

\begin{figure}[t]
\centerline{\includegraphics[scale=0.28]{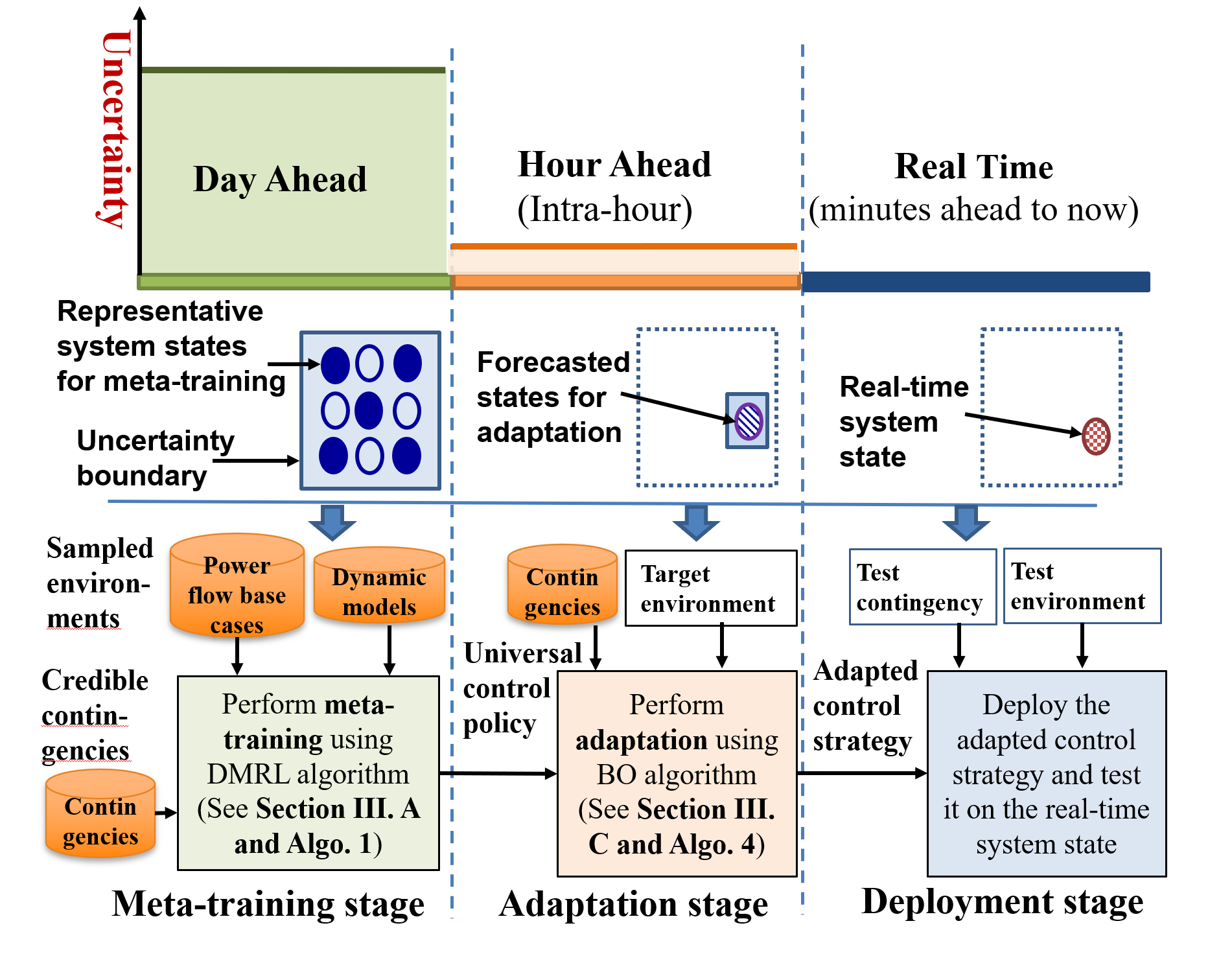}}
\caption{An illustration of application of the proposed method for grid control and its fitting into existing grid operation stages (or time frames). (Note: While all the states are shown within the forecasted operation boundary here, which is valid most of the time, we acknowledge that there are some rare situations where the system operation state at the adaptation stage or deployment stage might be out of the boundary considered during the meta-training stage. We have tested such rare conditions in Section \ref{sec:results}.)}
\label{fig:metalearning4grid}
\end{figure}

We integrate the MSO idea with the highly-scalable PARS algorithm to develop a novel DMRL algorithm that can quickly adapt to the new target power grid environment $\mathcal{E}_{tar}$ (herein a new power grid environment can be either a new power flow condition, or new set of system dynamic parameters or a combination of both). The key idea is illustrated in Fig. \ref{fig:metalearning4grid}. The DMRL algorithm includes three stages:
\begin{enumerate}
\item meta-training;
\item adaptation;
\item real-time deployment.
\end{enumerate}

%{\color{blue}Fig.\ref{fig:dmrl_flowchart} show the flowchart and connections of the meta-learning, adaptation, and testing stages of the proposed DMRL method.}

At the \textbf{day-ahead meta-training stage}, our work is based on the fact that grid operators do not have the exact information of load and generation patterns, load dynamics, and the exact fault locations and durations, and that they usually have some prior knowledge about system uncertainties based on historical operation data in the day-ahead operation as illustrated in Fig. \ref{fig:uncertainties_grid}. Therefore,
we propose to train a meta-learning model based on a set of representative forecasted system environments $\lbrace \bm{\mathcal{E}}_{trn}\rbrace$. The meta-learning stage produces a well-trained universal policy $\pi_{\theta}(s,c)$ that is conditioned on and can work for a distribution of different environments $\bm{P}(\lbrace \bm{\mathcal{E}}_{trn}\rbrace)$. Note that if the system operation conditions do not change notably from day to day, this meta-training stage is not required to perform every day, because the universal policy model can be reused. 

At the \textbf{hour-ahead (or intra-hour) adaptation stage}, within a short period of time (from 5 minutes to 1 hour) before actual real-time operation, improved forecast of the power grid states with much less uncertainty is obtained and thus the target environment $\mathcal{E}_{tar}$ can be constructed. For fast adaptation, we directly use the well-trained universal policy $\pi_{\theta}(s,c)$ from the meta-learning stage, and optimize the latent vector $c(\mathcal{E}_{tar})$ to adapt it to the target environment $\mathcal{E}_{tar}$. Consequently, an adapted control strategy $\pi_{\theta}(s,c(\mathcal{E}_{tar}))$ (i.e, an optimized universal policy combined with the optimized latent vector) for the target grid environment is created. With this proposed procedure of meta-learning and adaptation, much less time will be required during the adaptation stage for a target or new environment. Thus, it can meet the real-time operation requirements.

At the \textbf{deployment stage}, the adapted DMRL-based control strategy will be tested for the real-time emergency control. The testing environment could still have a small (i.e., 1\% to 2\%  short-term forecasting errors \cite{li2020novel}) difference from the specific target environment we consider at the adaptation stage. As we will shown in Section \ref{sec:results}, the generalization capability of the DMRL model can accommodate and handle such small difference well.

In terms of fitting into real-world power grid operation procedures, the three stages discussed above directly correspond to day-ahead, hour-ahead, and real-time operation time frames, and improve forecasting over time. 

% end of blue for the added subsection to show the flow chart of key process

\section{Deep meta reinforcement learning}\label{sec:dmrl}

In this section, we present the key algorithms and implementation details of our proposed DMRL method.
Fig. \ref{fig:dmrl_flowchart} shows the flowchart and connections of the meta-learning, adaptation, and testing stages of the proposed DMRL method.
\begin{figure}[t]
\centerline{\includegraphics[scale=0.28]{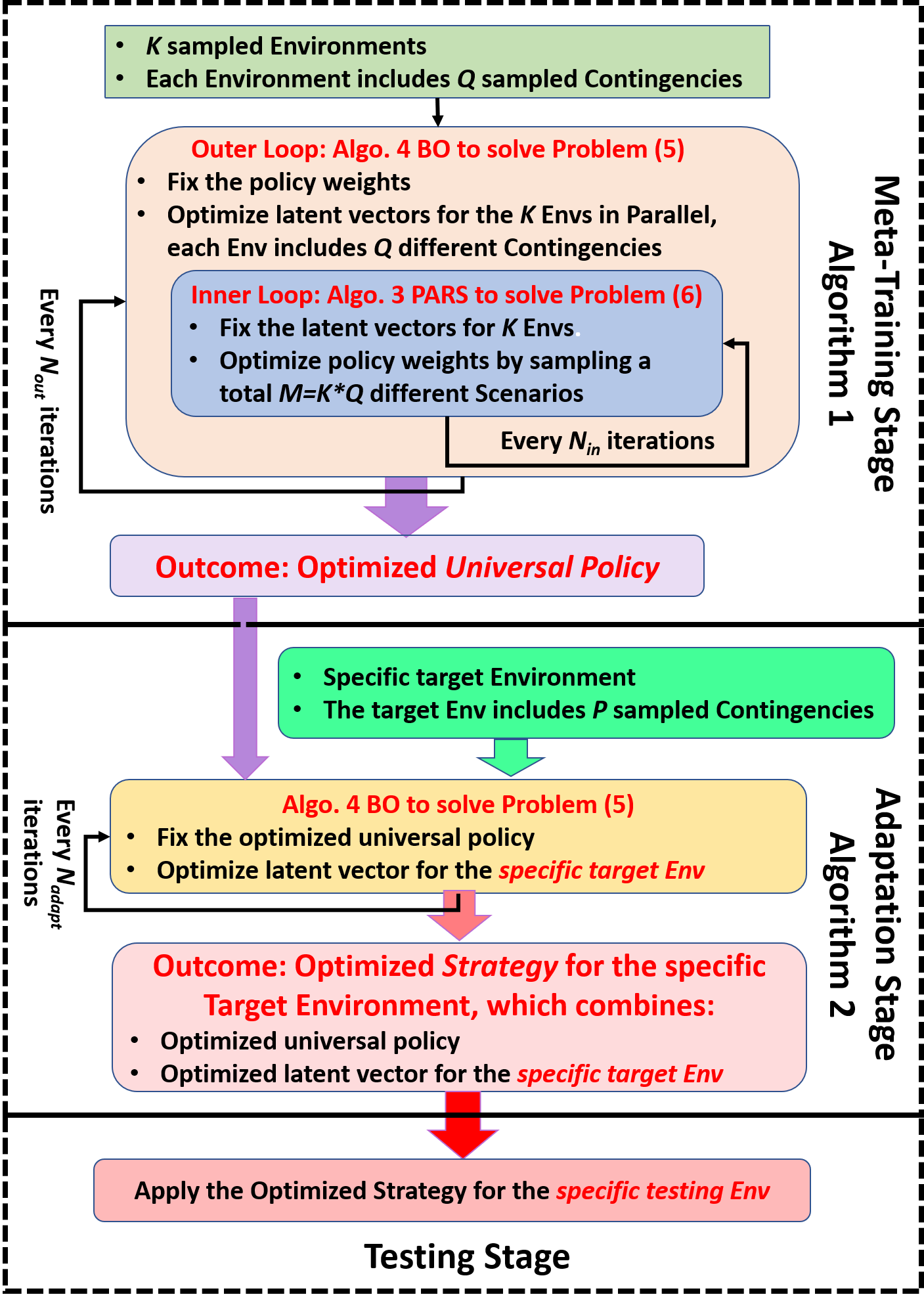}}
\caption{Flow Chart of the three stages of the proposed DMRL Method}
\label{fig:dmrl_flowchart}
\vspace{-10pt}
\end{figure}

\subsection{Deep Meta-Reinforcement Learning through Meta-Strategy Optimization}

We propose a novel DMRL algorithm that combines the MSO and PARS algorithms to solve the meta-learning problem defined in (\ref{eqn:metarl_obj}). Our proposed method learns a latent-vector-conditioned policy \comment{$\pi_\theta(s, c)$}on a distribution of Environments $\bm{P}(E_{i})$ representing forecasted grid operation conditions (different power flows and system dynamic parameters), and can quickly adapt the trained policy to a new target environment. Assuming that the distribution of Environments $\bm{P}(E_{i})$ defined in Section \ref{sec:prob_formul} could be represented by a latent space $\mu$, as explained in Section II.C, MSO extends the standard RL policy learning by training a universal policy $\pi_{\theta}(s,c)$ that is conditioned on the latent vector $c$ in the latent space $\mu$, $c\in\mu$. We can train such a universal policy with any standard RL algorithm by treating the latent vector $c$ as part of the observations \cite{yu2017UCP}. The trained universal policy will change its behaviors with respect to different latent vectors in the latent space, thus a policy with a particular latent vector input $c\in\mu$ can be treated as one control \textit{strategy}, which is trained to be optimal for one task in $\bm{P}(E_{i})$ as defined in Section \ref{sec:prob_formul}.

At the meta-training stage, we target at learning a universal policy $\pi_{\theta}(s,c)$ conditioned on a distribution of environments $\bm{P}(E_{i})$. We need to determine the corresponding latent vectors in the latent space, ${\bm{c}\in\mu}$, for a set of training environments $\bm{\mathcal{E}}\in\bm{P}(E_i)$. Given that our goal is to find a set of strategies that work well for the set of training environments $\bm{\mathcal{E}}\in\bm{P}(E_i)$, a direct approach is to use the performance of the training environments, i.e., the accumulated reward of the training environment as the metric to search the optimal strategies. Mathematically, we reformulate (\ref{eqn:metarl_obj}) and solve the following optimization problem on a distribution of environments $\bm{P}(E_{i})$ at the meta-training stage:
\begin{equation}
\theta^\ast=\arg\max _{\theta} \mathbb{E}_{\bm{\mathcal{E}}\in\bm{P}(E_{i})} \left[ \max _{c\in\mu}{J}\left(c,\theta\right)\right],
\label{eqn:mso_1}
\end{equation}
where $\theta$ is the weight of the policy network, $J\left(c,\theta\right)$ is the expected accumulative reward of the strategy $\pi_\theta(s,c)$ on the environment $\mathcal{E}\in\bm{P}(E_i)$, and $c$ is the latent vector that encodes the power grid operation condition environment $\mathcal{E}$ in the low-dimensional latent space $\mu$, $c\in\mu$.
It is rather challenging to directly solve (\ref{eqn:mso_1}) because the optimization of $J(c, \theta)$ is inside the expectation of the objective term, and every single evaluation of $\theta$ requires solving a optimization problem to find the optimized latent vector $c$ to maximize $J(c, \theta)$, which is computationally expensive. To overcome this, we revise (\ref{eqn:mso_1}) as:
\begin{equation}
\theta^\ast,c\left(\mathcal{E}\right)^\ast=\arg\max _{\theta, c(\mathcal{E})} \mathbb{E}_{\bm{\mathcal{E}}\in\bm{P}(E_{i})} \left[{J}\left(c(\mathcal{E}),\theta\right)\right],
\label{eqn:mso_2}
\end{equation}
where $c(\mathcal{E})$ is the specific latent vector that encodes the specific training environment $\mathcal{E}$ in the low-dimensional latent space $\mu$, $\mathcal{E}\in\bm{P}(E_i)$ and $c(\mathcal{E})\in\mu$. Then, we can solve the optimization problem in (\ref{eqn:mso_2}) by solving two related optimization problems defined in (\ref{eqn:mso_3}) and (\ref{eqn:mso_4}) in an alternative manner, where $k$ is the iteration number.
\begin{equation}
c(\mathcal{E})_{k+1}=\arg{\max_{c}{J}}\left(c,\theta_k\right), \forall \mathcal{E}\in\bm{P}(E_{i})
\label{eqn:mso_3}
\end{equation}
\vspace{-3mm}
\begin{equation}
\theta_{k+1}=\arg\max _\theta \mathbb{E}_{\mathcal{E}\in\bm{P}(E_{i})} \left[J\left(c(\mathcal{E})_{k},\theta\right)\right]
\label{eqn:mso_4}
\end{equation}

Note that (\ref{eqn:mso_3}) defines the \textit{strategy optimization} (SO) problem to optimize the latent vector $c(\mathcal{E})$ for a strategy $\pi_\theta(s,c)$ and it will be solved by the Bayesian optimization (BO) described in Subsection III.C, and (\ref{eqn:mso_4}) defines a RL problem to determine the policy network parameters $\theta$ and could be solved by DRL algorithms. We solve (\ref{eqn:mso_4}) using PARS algorithm presented in the following subsection.  \textbf{Algorithm 1} gives the details of the proposed DMRL algorithm solving (\ref{eqn:mso_3}) and (\ref{eqn:mso_4}).  As shown in \textbf{Algorithm 1} and Fig. \ref{fig:dmrl_flowchart}, the meta-learning algorithm includes two loops, the outer BO loop and the inner PARS loop. In the outer BO loop, for each iteration, we keep the weights $\theta$ of the universal policy $\pi_{\theta}(s,c)$ fixed and sample $K$ different environments $\mathcal{E}_{i}|i = 1,2,..,K$. We use Bayesian optimization to update the optimal latent vector $c(\mathcal{E}_{i})$ for each corresponding sampled environment $\mathcal{E}_{i}$ by solving (\ref{eqn:mso_3}). Note that during the Bayesian optimization to update the optimal latent vector $c(\mathcal{E}_i)$, for each different environments $\mathcal{E}_{i}$, we sample a total of $Q$ different contingencies. In the inner PARS loop, for each iteration, we sample $M$ different scenarios $\Gamma_{m\in M}=\{\mathcal{E}_{i\in K}\cup\mathcal{C}_{j\in Q}\}$, which is the combination of $K$ different environments $\mathcal{E}$ and $Q$ different contingencies $\mathcal{C}$. We keep the latent vector $c(\mathcal{E}_{i})$ for each sampled environment $\mathcal{E}_i$ fixed and use PARS to update the weights $\theta$ of the universal policy $\pi_{\theta}(s,c)$ by solving (\ref{eqn:mso_4}). At the end, this stage generates a  well-trained universal policy $\pi_{\theta}(s,c)$ that is conditioned on the optimized latent vectors and works well for the sampled environments from the distribution of the environments $\bm{P}(E_{i})$.

At the adaptation stage, we target at transferring the universal policy $\pi_{\theta}(s,c)$ trained with DMRL in \textbf{Algorithm 1} at the meta-learning stage to a target environment $\mathcal{E}_{tar}\in\bm{P}(E_{i})$. We need to determine the corresponding latent vector $c(\mathcal{E}_{tar})$ of the strategy $\pi_{\theta}(s,c)$ for the new specific target environment $\mathcal{E}_{tar}$. Mathematically, at the adaptation stage, we  need to solve the strategy optimization (SO) problem defined in (\ref{eqn:mso_3}) for only a specific target environment $\mathcal{E}_{tar}$, which could achieve high computational efficiency. \textbf{Algorithm 2} gives the details of the proposed adaptation algorithm to optimize the specific latent vector $c(\mathcal{E}_{tar})$ for the target environment $\mathcal{E}_{tar}$. As shown in \textbf{Algorithm 2} and Fig. \ref{fig:dmrl_flowchart}, we sample $P$ different  contingencies {$\mathcal{C}_j|j = 1,2,..,P$} and  optimize  the  latent  vector $c(\mathcal{E}_{tar})$ against the $P$ different  contingencies  for  this  specific  target environment $\mathcal{E}_{tar}$. Note also that in the proposed DMRL method, we adopt the same strategy optimization process to optimize the latent vector for the policy at both the meta-training and adaptation stages prior to testing/deployment. The outcome of the adaptation stage is a specific optimized strategy (an optimized universal policy $\pi_{\theta}(s,c)$ together with the optimized specific latent vector $c(\mathcal{E}_{tar})$) for the specific target environment $\mathcal{E}_{tar}$, and this optimized strategy is robust to a wide range of different contingencies.
 % end of blue color

\vspace{-3pt}
\begin{algorithm}
\label{algo:mso}
 \caption{Deep Meta-Reinforcement Learning (DMRL)}
 \begin{algorithmic}[1]
  \STATE initialize policy weights {$\theta_0$} with random small numbers
  \STATE initialize latent vectors {$\bm{c}(\mathcal{E})_{0}$} as $\bm{0}$
  \FOR {{outer\_iter} $k$ = 0 \TO $N_{out}$}
    \STATE sample $K$ \textit{Environments} \{$\mathcal{E}_{i}|i = 1,2,..,K$\}
    \STATE for each {$\mathcal{E}_{i}$}, sample $Q$ \textit{Contingencies} {$\mathcal{C}_j|j = 1,2,..,Q$}, solve (\ref{eqn:mso_3}) using Bayesian optimization with fixed $\theta_k$ and update corresponding latent vector {$c(\mathcal{E}_i)_{k}$}
    \FOR {{inner\_iter} $q$ = 0 \TO $N_{in}$}
    \STATE sample $M$ \textit{Scenarios} $\Gamma_{m\in M}=\{\mathcal{E}_{i\in K}\cup\mathcal{C}_{j\in Q}\}$
    \STATE {update $\theta_{k}$} by solving (\ref{eqn:mso_4}) with the PARS algorithm with fixed {$\bm{c}(\mathcal{E})_{k}$} as part of the observation
    \ENDFOR
  \ENDFOR
 \RETURN {$\theta_k $}, {$\bm{c}(\mathcal{E})_{k}$}
\end{algorithmic}
\end{algorithm}
\vspace{-3pt}

\vspace{-3pt}
\begin{algorithm}
\label{algo:mso_adaptation}
 \caption{Adaptation}
 \begin{algorithmic}[1]
  \STATE define the specific target environment $\mathcal{E}_{tar}$ for adaptation
  \STATE load policy weights {$\theta_{train}$} from the outcome of \textbf{Algorithm 1}
  \STATE initialize the specific latent vector {$c(\mathcal{E}_{tar})$} for the target environment $\mathcal{E}_{tar}$ as $\bm{0}$
  \FOR {{adapt\_iter} $i$ = 0 \TO $N_{adapt}$}
     \STATE sample $P$ \textit{Contingencies} {$\mathcal{C}_j|j = 1,2,..,P$}
     \STATE solve (\ref{eqn:mso_3}) using Bayesian optimization with fixed $\theta_{train}$ and update the specific latent vector {$c(\mathcal{E}_{tar})$}
  \ENDFOR
 \RETURN the optimized specific latent vector {$c(\mathcal{E}_{tar})$}
\end{algorithmic}
\end{algorithm}
\vspace{-3pt}

\subsection{Parallel Augmented Random Search}

 Augmented random search(ARS) is a derivative-free, efficient, easy-to-tune and robust-to-train RL algorithm\cite{mania2018_ARS}. In our previous work\cite{huang2020accelerated_DRL}, we developed the PARS algorithm to scale up the ARS algorithm for large-scale grid control problems and reduce the training time. Specifically, we proposed and and implemented a novel nest parallelism scheme for ARS on a high-performance computing platform. The key steps of PARS algorithm are shown in \textbf{Algorithm 3}, with more details presented in \cite{huang2020accelerated_DRL}. We use it to solve the RL problem defined with (\ref{eqn:mso_4}) in the DMRL \textbf{Algorithm 1}.

\vspace{-3pt}
\begin{algorithm}
 \caption{Parallel Augmented Random Search (PARS)}
 \begin{algorithmic}[1]
  \STATE Initialize: Policy weights {$\theta_0$}, the running mean of observation states {$\mu_0 = \mathbf{0} \in \mathbb{R}^n$} and the running standard deviation of observation states {$\Sigma_0 = \mathbf{I}_{n} \in \mathbb{R}^n$}, the total iteration number $H$.
  \FOR {iteration $t = 1,...,H$}
    \STATE sample $N$ random directions {$\delta_1, ..., \delta_N$} with the same dimension as policy weights $\theta$
    \FOR {each $\delta_i  (i \in [1,..., N])$}
    \STATE add $\pm$ perturbations to policy weights: $\theta_{ti+} = \theta_{t-1} + \upsilon \delta_i$ and $\theta_{ti-} = \theta_{t-1} - \upsilon \delta_i$
    \STATE \textbf{do} total $2M$ \textit{Scenarios} (episodes) denoted by $R_{\Gamma_{m \in M}}$, calculate the average rewards for $\pm$ perturbations
    % {
    %     \vspace{-3mm}
    %     \begin{equation}{
    %     \begin{cases}
    %         \text{$r_{ti+} = \frac{1}{m}R_{p \in T}(\theta_{ti+} , \mu_{t-1} , \Sigma_{t-1})$} \\
    %         \text{$r_{ti-} = \frac{1}{m}R_{p \in T}(\theta_{ti-}, \mu_{t-1} , \Sigma_{t-1})$}\\
    %     \end{cases}}
    %     \end{equation}
    %     \vspace{-3mm}
    % }

    \STATE During each episode, normalize states $s_{t,k}$ and obtain the action $a_{t,k}$, and new states $s_{t,k+1}$. Update $\mu_t$ and $\Sigma_{t}$ with  $s_{t,k+1}$
    % {
    % \vspace{-3mm}
    %     \begin{equation}{
    %     \begin{cases}
    %         \text{$s_{t,k} = (s_{t,k} - \mu_{t-1})/\Sigma_{t-1}$} \\
    %         \text{$a_{t,k} = \pi_{\theta_t}(s_{t,k})$} \\
    %         \text{$s_{t,k+1} \longleftarrow \mathcal{P}(s_{t,k},a_{t,k})$} \\
    %     \end{cases}}
    %     \label{eq: forward_pass}
    %     \end{equation}
    % \vspace{-3mm}
    % }
    \ENDFOR
    \STATE sort the directions based on $\max[r_{ti+},r_{ti-}]$ and select top $b$ directions, calculate their standard deviation $\sigma_b$
    \STATE update the policy weight:
    {
    \vspace{-3mm}
    \begin{equation}\label{eq: weights_update}
    \theta_{t+1} = \theta_t + \frac{\alpha}{b\sigma_b}\sum\limits_{i=1}^{b}(r_{ti+}-r_{ti-})\delta_i
    \end{equation}
    \vspace{-3mm}
    }
    \STATE Step size $\alpha$ and standard deviation of the exploration noise $\upsilon$ decay with rate $\varepsilon$: $\alpha = \varepsilon\alpha$, $\upsilon = \varepsilon\upsilon$
  \ENDFOR
 \RETURN {$\theta $}
\end{algorithmic}
\label{ARS_algo}
\end{algorithm}

\comment {\vspace{-18pt}}
\subsection{Strategy Optimization through Bayesian Optimization}

We propose to use Bayesian optimization (BO)\cite{mockus2012bayesian} to solve the SO problem defined with (\ref{eqn:mso_3}) in the DMRL \textbf{Algorithm 1}, which can handle noisy and often non-convex objectives. BO is a parameter optimization method for any black-box objective function \textit{f}(\textit{x}). The core idea of BO is to conduct sequential sampling to construct the target function.
BO consists of two key elements, a probabilistic surrogate model, which is a distribution over the target function, and an acquisition function, which is used to explore the parameter space based on the surrogate model.
The surrogate model \textit{P}(\textit{f}) can be expressed as follows:
\begin{equation}
P(f|{x_{1:t}},{y_{1:t}}) = \frac{{P({x_t},{y_t}|f)P(f|{x_{1:t - 1}},{y_{1:t - 1}})}}{{P({x_t},{y_t})}}
\label{eqn:prob_model}
\end{equation}
\noindent where $x_{1:t}$ is a matrix containing all the inputs from time step 1 to $t$, and $y_{1:t}$ is a vector containing the corresponding outputs, i.e., $y_{t}$ = $f(x_{t})$.
The next input $x_{t+1}$ is selected by minimizing the following expected loss:
\begin{equation}
{x_{t + 1}} = \mathop {\arg \min }\limits_{{x_{t + 1}}} \int {{\delta _t}(f,{x_{t + 1}})dP(f|{x_{1:t}},{y_{1:t}})}
\label{eqn:exp_utilityl}
\end{equation}
where $\delta_{t}(f,x_{t+1})$ is called the regret function, which measures the difference between the sampling point $x_{t+1}$ and the optimal point $x^{*}$.
%Some exemplary forms of the regret function include the optimality gap:$f(x_{t+1})–f(x^{*})$ and the Euclidean distance: $||x_{t+1} – x^{*}||^{2}$.

However, in reality, since we do not have the knowledge of the optimum $x^{*}$, (\ref{eqn:exp_utilityl}) cannot be applied directly. An acquisition function $\alpha_{t}(x)$ is designed instead as a proxy of the regret function. Namely,  (\ref{eqn:exp_utilityl}) becomes:
\begin{equation}
{x_{t + 1}} = \arg {\min _{x_{t + 1}}}{\alpha _t}(x)
\label{eqn:acql}
\end{equation}
Some options for the acquisition function include probability of improvement (PI), expected improvement (EI), and upper confidence bound (UCB). In this work, we use UCB as the acquisition function, which has the following expression:\\
\begin{equation}
{x_{t + 1}} = \arg {\max _{x_{t + 1}}}(\mu_{t}(x) + \kappa * \sigma_{t}(x))
\label{eqn:ucb}
\end{equation}
In (\ref{eqn:ucb}), $\mu_{t}(x)$ and $\sigma_{t}(x)$ are the mean and standard deviation of a Gaussian Process(GP), and $\kappa$ is the weight factor.

\comment {UCB can be seen as the maximum value of the weighted sum of the expected performance and the uncertainty over all the solutions. $\kappa$ is the weight factor. A smaller $\kappa$ implies more exploration while a larger $\kappa$ indicates that the algorithm tends to select more stable solutions. }

The overall process of applying BO to solve (\ref{eqn:mso_3}) is shown in \textbf{Algorithm 4}. Note that for (\ref{eqn:mso_3}), the BO objective function $f(x)$ is the expected accumulative reward $J(c, \theta)$ with only $c$ as the optimization variable $x$, since the $\theta$ is fixed for (\ref{eqn:mso_3}).
\comment {The initial probabilistic surrogate model is built upon random sampling. During the optimization, the next data point $x_{t+1}$ is sampled by minimizing the acquisition function, and the corresponding output $y_{t+1}$ is calculated by the true target function $f$. The surrogate model $P(f)$ is updated based on the new data point.}
\begin{algorithm}
\label{algo:bo}
 \caption{Bayesian Optimization}
 \begin{algorithmic}[1]
  \STATE initialize the sampling set $D$, probabilistic surrogate model $P(J(c,\theta_{fix}))$, solution $c$, and acquisition function $\alpha _t(c;D)$
  \FOR {$t=1,2,..., N_{BO}$}
     \STATE select ${c_{t + 1}} = \arg {\min _{c_{t + 1}}}{\alpha _t}(c;D)$
    \STATE calculate $y_{t+1} = J(c_{t+1},\theta_{fix})$
    \STATE update the sampling set $D=D\cup(c_{t+1},y_{t+1})$
    \STATE update $P(J(c,\theta_{fix}))$ and $\alpha _t(c_{t+1};D)$
  \ENDFOR
  \STATE return the solution $c$
\end{algorithmic}
\end{algorithm}

\comment{\subsection{Flowchart of the Proposed DMRL Method and Discussion on How It Naturally Fits the Real Grid Operation}

Fig.\ref{fig:dmrl_flowchart} illustrates the key concepts and connections of the meta-learning, adaptation, and testing stages of the proposed DMRL method, as well as how the proposed DMRL method naturally fits into real power grid operation procedures as follows.

At the \textbf{day-ahead meta-learning stage}, our work is based on the fact that grid operators do not have the exact information of load and generation patterns, load dynamics and the exact fault locations and durations and that the grid operators usually have good knowledge about system uncertainties in day-ahead operation as illustrated in Fig. \ref{fig:uncertainties_grid}. Therefore, we first train a universal policy on a distribution of environments that can be derived from the current operation conditions and uncertainties, and we need also to determine the corresponding latent vectors in the latent space for each sampled individual environment in the set of training environments. Mathematically, in the meta-learning stage, we need to alternatively solve problem (\ref{eqn:mso_3}) with BO algorithm and problem (\ref{eqn:mso_4}) by PARS algorithm. In the detailed implementation, during this meta-learning stage, we sample $K$ different environments, and for each environment we sample $Q$ different contingencies. The outcome of the meta-learning stage is the well-trained universal policy that could work for a distribution of different environments.

At the \textbf{intra-hour adaptation stage}, we assume we have a much better estimation of the load/generation pattern and of the load dynamics, but still not on the fault location nor on its duration. At this adaptation stage, we optimize only the specific latent vector for a specific target environment that is defined based on the much better estimation of the load/generation pattern and of the load dynamics of the grid, which is different from the environments we sampled during the meta-learning stage. We directly use the well-trained universal policy from the meta-learning stage. Mathematically, in the meta-adaptation stage, we need only to solve problem (\ref{eqn:mso_3}) by BO. In the detailed implementation, during this adaptation stage, we define one specific target environment, we sample $P$ different contingencies and optimize the latent vector against the $P$ different contingencies for this specific target environment. We also assume that, due to unpredicted and extreme situations, the load/generation pattern and the load dynamics of this specific target environment at the adaptation stage can be out of the boundary/range of the environments we used for the meta-learning. The outcome of the adaptation stage is a specific optimized strategy (an optimized universal policy together with the optimized specific latent vector) for the specific target environment, and this optimized strategy is robust to a wide range of different contingencies.

At the \textbf{real-time testing stage}, we assume the testing environment still has a small (around 2.1\%) difference when compared with the specific target environment we defined at the adaptation stage, this is because even the state-of-art short term load forecasting technique \cite{li2020novel} still have a small forecasting error. We just deploy the optimized strategy obtained from the adaptation stage for the testing environment at the testing stage.

\begin{figure}[t]
\centerline{\includegraphics[scale=0.36]{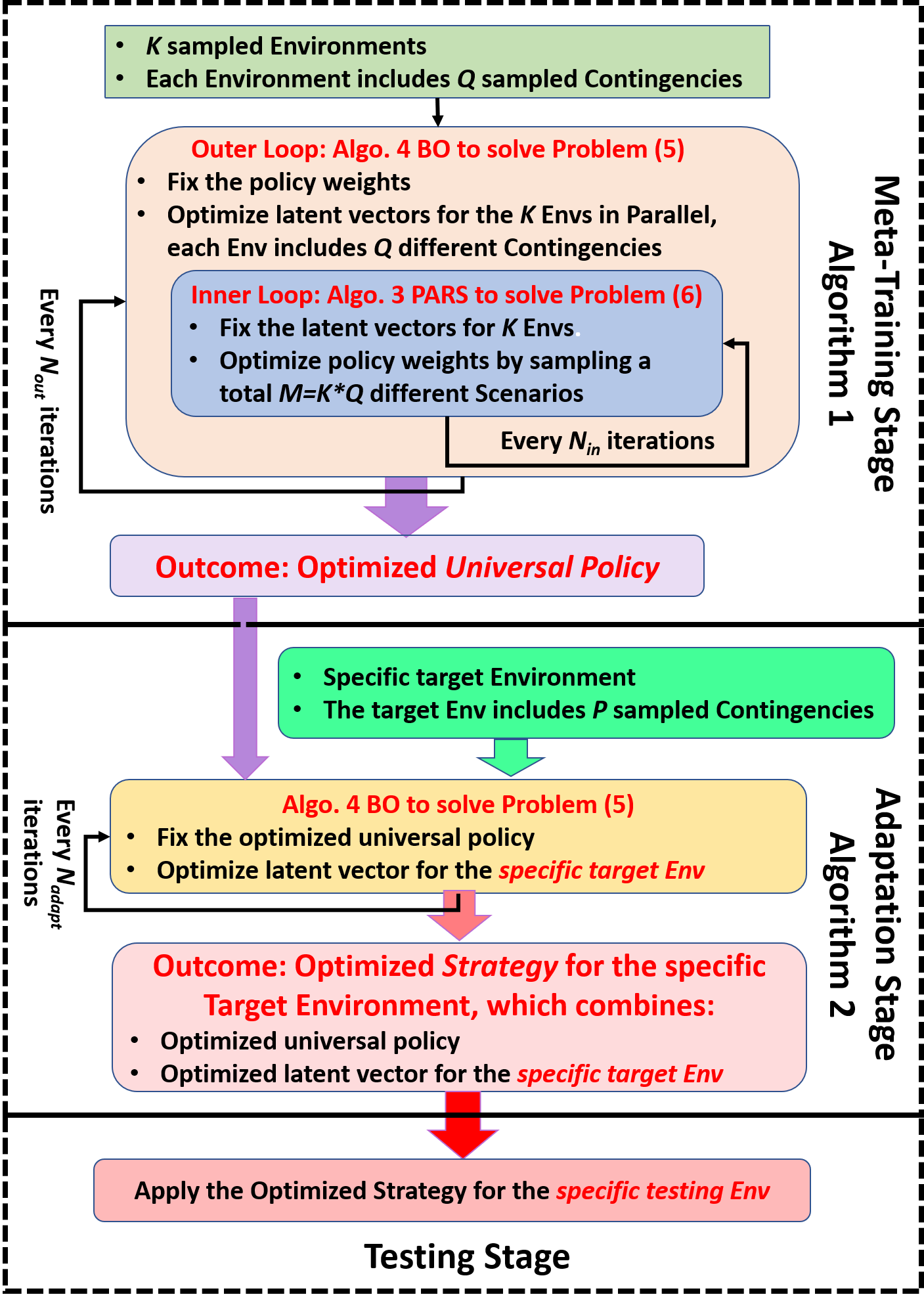}}
\caption{Flow Chart of the three stages of the proposed DMRL Method}
\label{fig:dmrl_flowchart}
\vspace{-10pt}
\end{figure}

% end of blue for the added subsection to show the flow chart of key process
}% end of comment of this subsection

\comment {Note that in SO, the processes of obtaining $\mu$ at the training and adaptation stages are different: one is through a projection network, while the other is through optimization methods . This mismatch between training and adaptation phases leads to a learned latent space that is not suitable for application scenarios that require fast adaptation.

\textit{In the MSO method, we adopt the same BO-based adaptation process to obtain the latent input $\mu$ to the policy during both meta-training and adaptation prior to testing or deployment.} As such, it learns a latent vector conditioned policy $\pi_\theta(s,\mu)$ on a large variety of environments representing forecasted grid operation conditions and/or a list of critical emergency events, and can quickly adapt the trained policy to a new target environment. }

\comment {In this algorithm, the policy network is a feed-forward neural network with two hidden layers. The weights of the network, $\theta$, is the goal to be optimized by augmented random search (ARS). To adapt the policy to each different environments, a latent vector, $h_i$, need to be learned to represent environment $i$, which in this study, a distinct fault case $c_i$. The latent vector $h$ has a dimension of $(2, )$ in this study. To begin the algorithm, $\theta$ is first initialized with random small numbers, and each $h_i$ is initialized as $(0.0, 0.0)$. This algorithm has two nested for loops: in each outer for loop, the latent vectors of a subset of fault cases are updated through Bayesian optimization, in which a Gaussian process is used to model $R_c \gets f(h_c)$ under the fixed $\theta$. In each inner for loop, $\theta$ is then optimized through ARS. Specifically, a number of $k$ very small perturbations $\delta_i$, $ i\gets 0: k$ are made to the policy weight $\theta$ in both positive and negative ways, and the corresponding rollout rewards  $r_{i+}\gets\sum\limits_{c}{R_c(h_c, \theta+\delta_i)}$ and $r_{i-}\gets\sum\limits_{c}{R_c(h_c, \theta-\delta_i)}$. The top $m$ perturbations with best performing rewards are selected and  averaged with weights of the effectiveness of perturbation $(r_{i+}-r_{i-})$. The policy weight $\theta$ then will be updated by this averaged perturbation with step size $\alpha$.
\begin{equation}
    \theta_{t+1} = \theta_t + \frac{\alpha}{m}\sum\limits_{i=0}^{m}(r_{i+}-r_{i-})\delta_i
\end{equation}
}
\vspace{-6pt}
\comment{
\subsection{Adaptation Process}
{\color{red}[Renke]Briefly introduce the adaptation process}
}

%emphasize difference from existing Meta RL, like RL^2, MAML

\section{Test cases and Results}\label{sec:results}

\subsection{The FIDVR problem and MDP formulation}
\begin{figure}
\centering
\includegraphics[width=0.42\textwidth]{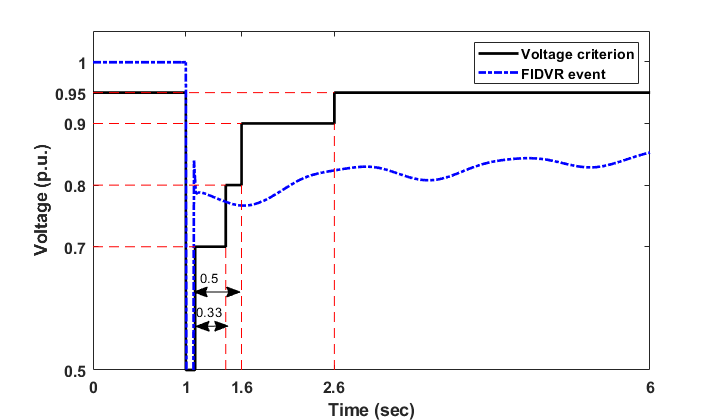}
\caption{Transient voltage recovery criterion for transmission system \cite{PJM2009}}
\label{fig:volt_performance_curve}
\end{figure}

Fault-induced delayed voltage recovery (FIDVR) is defined as the phenomenon whereby system voltage remains at significantly reduced levels for several seconds after a fault has been cleared \cite{NERC2009}. The root cause is stalling of air-conditioner (A/C) motors and prolonged tripping. FIDVR events occurred in many utilities in the US. The industry have concerns over FIDVR issues since residential A/C penetration is at an all-time high and continues to grow. A transient voltage recovery criterion is defined to evaluate the system voltage recovery. Without loss of generality, we referred to the standard shown in Fig. \ref{fig:volt_performance_curve} \cite{PJM2009}. The objective of emergency control for FIDVR problem is to shed as little load as possible to recover voltages to meet the voltage recovery criterion.

Key elements of the MDP formulation of this problem are:
\begin{enumerate}
    \item \textit{Environment}: The grid environment is simulated with RLGC \cite{Huang2020_DRL, RLGC}. We considered a modified IEEE 300-bus system with loads larger than 50 MW within Zone 1 represented by WECC composite load model {\cite{huang2019_cmpldw}}. We prepared a wide range of power flow cases and considered different combinations of  dynamic load parameters that are key for FIDVR problems for both training and testing.
    \item \textit{Action}: Load shedding control actions are considered for all buses with dynamic composite load model at Zone 1 (46 buses in total). The percentage of load shedding at each control step could be from 0 to 20\%. The agent is designed to provide control decisions (including no action as an option) to the grid every 0.1 second.
    \item \textit{Observation}: The observations included voltage magnitudes at 154 buses within Zone 1, as well as the remaining fractions of 46 composite loads. Thus the dimension of the observation space is 200. The agent needs to obtain the observations from the grid environment every 0.1 second.
    \item \textit{Reward}: the reward $r_t$ at time $t$ is defined as follows:
\begin{equation}\label{eq: UVLS_reward}
 r_t= 
\begin{cases}
    -10000, \text{ if } V_i(t)<0.95, \quad t>T_{pf} +4		 \\
    c_1 \sum_i \Delta V_i(t) -c_2 \sum_j \Delta P_j (t) -c_3u_{iv},  \text{ otherwise}
\end{cases}
\end{equation}
\[
\Delta V_i(t)= 
\begin{cases}
    min \left\lbrace V_i(t) - 0.7, 0 \right\rbrace , \text{ if }  {\scriptstyle T_{pf}<t<T_{pf} +0.33	}	 \\
    min \left\lbrace V_i(t) - 0.8, 0 \right\rbrace , \text{ if }  {\scriptstyle T_{pf} + 0.33<t<T_{pf} +0.5	}\\
    min \left\lbrace V_i(t) - 0.9, 0 \right\rbrace , \text{ if }  {\scriptstyle T_{pf} + 0.5<t<T_{pf} +1.5}	\\
    min \left\lbrace V_i(t) - 0.95, 0 \right\rbrace , \text{ if } {\scriptstyle  T_{pf} + 1.5<t}
\end{cases}
\]
where $T_{pf}$ is the time instant of fault clearance, $V_i(t)$ is the bus voltage magnitude for bus $i$ in the power grid, $\Delta P_j (t)$ is the load shedding amount in p.u. at time step $t$ for load bus $j$, and $u_{iv}$ is the invalid action penalty. $c_1, c_2,$ and $c_3$ are weight factors. 
   
    \item Other details such as the state transition can be found in \cite{Huang2020_DRL}.
    
\end{enumerate}

\begin{table*}[t]
\caption{Power flow conditions for training }
\begin{center}
\begin{tabular}{|c|c|c|}
\hline
\textbf{Training power flow case } & \textbf{Generation} & \textbf{Load} \\
\hline
{\textbf{1}} & Total 22929.5 MW (100\%) & Total 22570.2 MW (100\%)\\ 
\hline
{\textbf{2}}& 120\% for all generators & 120\% for all loads \\ 
\hline
{\textbf{3}}& 135\% for all generators & 135\% for all loads \\ 
\hline
{\textbf{4}}& 115\% for all generators & 150\% for loads in Zone 1 \\ 
\hline

\end{tabular}
\label{tab_pfcases_training}
\end{center}
\end{table*}

\begin{table*}[t]
\caption{Power flow conditions for adaptation and testing}
\begin{center}
\begin{tabular}{|c|c|c|c|c|}

\hline

& \multicolumn{2}{c|}{ \textbf{Adaptation} } & \multicolumn{2}{c|}{ \textbf{ Testing} } \\ 
\cline{2-5}
\multirow{-2}{*}{\textbf{Adaptation/testing power flow case}} & {\textbf{Generation}} & 
{ \textbf{Load}} & 
{ \textbf{Generation}} & 
{ \textbf{Load} } \\ 
\hline

{\textbf{1}} & 90\% for all generators & 90\% for all loads & 92.4\% for all generators & 92.4\% for all loads\\ 
\hline
{\textbf{2}} & 110\% for all generators & 110\% for all loads & 107.7\% for all generators & 107.7\% for all loads\\ 
\hline
{\textbf{3}} & 115\% for all generators & 115\% for all loads & 117.2\% for all generators & 117.2\% for all loads\\ 
\hline
{\textbf{4}} & 125\% for all generators & 125\% for all loads & 122.5\% for all generators & 122.5\% for all loads\\ 
\hline
{\textbf{5}} & 140\% for all generators & 140\% for all loads & 142.1\% for all generators & 142.1\% for all loads\\ 
\hline
{\textbf{6}} & 95\% for all generators & 82.8\% for loads in Zone 1 & 97.1\% for all generators & 85.32\% for loads in Zone 1\\ 
\hline
{\textbf{7}} & 107\% for all generators & 124.4\% for loads in Zone 1& 104.6\% for all generators & 121.5\% for loads in Zone 1\\ 
\hline
{\textbf{8}} & 110\% for all generators & 134.3\% for loads in Zone 1& 112.3\% for all generators & 137.2\% for loads in Zone 1\\ 
\hline
{\textbf{9}} & 119\% for all generators & 159.7\% for loads in Zone 1& 121.1\% for all generators & 162.6\% for loads in Zone 1\\ 
\hline

\end{tabular}
\label{tab_pfcases_testing}
\end{center}
\end{table*}

\vspace{-12pt}
\begin{table}[t]
\caption{Dynamic models for training and testing}
\begin{center}
\begin{tabular}{|l|c|c|c|c|}
\hline
\textbf{ } & \textbf{A/C Motor Percentage} & $\bm{T_{stall}}(s)$ & $\bm{V_{stall}}(p.u.)$\\
\hline
{\textbf{Training}} & 44\%  & 0.032  & 0.45  \\ 

\hline
\multirow{4}*{ \textbf{Testing} } & 55\%  & 0.075  & 0.495 \\ 
\cline{2-4}
 & 55\%  & 0.1  & 0.53 \\
\cline{2-4}
 & 44\%  & 0.1  & 0.53 \\
\cline{2-4}
 & 35\%  & 0.05  & 0.45 \\

\hline
\end{tabular}
\label{tab_dyrcases}
\end{center}
\end{table}
\subsection{Performance metrics and baselines}

\begin{enumerate}
  \item \textit{Metrics for training and adaptation:} we target at day-ahead training (less than 24 hours) and hour-ahead policy adaptation (less than 60 minutes).

  \item \textit{Metrics for testing:} (a) the total solution time should be within 1 s for an event lasting 8 s to meet the real-time response requirement, or in other words, the average solution time should be within $0.0125$ s for each control step (80 steps in total); (b) the average reward on the testing scenarios set (the reward measures the optimality of the load shedding controls); (c) total load shedding amount. 

\end{enumerate}

We compared our method with PARS, one state-of-the-art DRL method, as well as the MPC method that provides (near) optimal solutions of the problem.

\subsection{Training, adaptation and testing of control strategies}

 The meta-learning, adaptation and testing were performed on a local high-performance computing cluster using 1152 cores. 
 
At the day-ahead meta-learning stage, since we do not know the exact load/generation pattern, as well as the load dynamics and the exact fault location and duration, we train a universal LSTM policy on 4 different environments and 18 different contingencies. Table \ref{tab_pfcases_training} defines the power flow conditions for training. We consider 18 different contingencies, which are a combination of 2 fault durations (0.05 and 0.08 s) and 9 candidate fault buses (2, 3, 5, 8, 12, 15, 17, 23, 26) in the contingency set $\lbrace C_{train}\rbrace$ for training.

\begin{figure}[t]
\centering
\includegraphics[width =0.8\columnwidth]{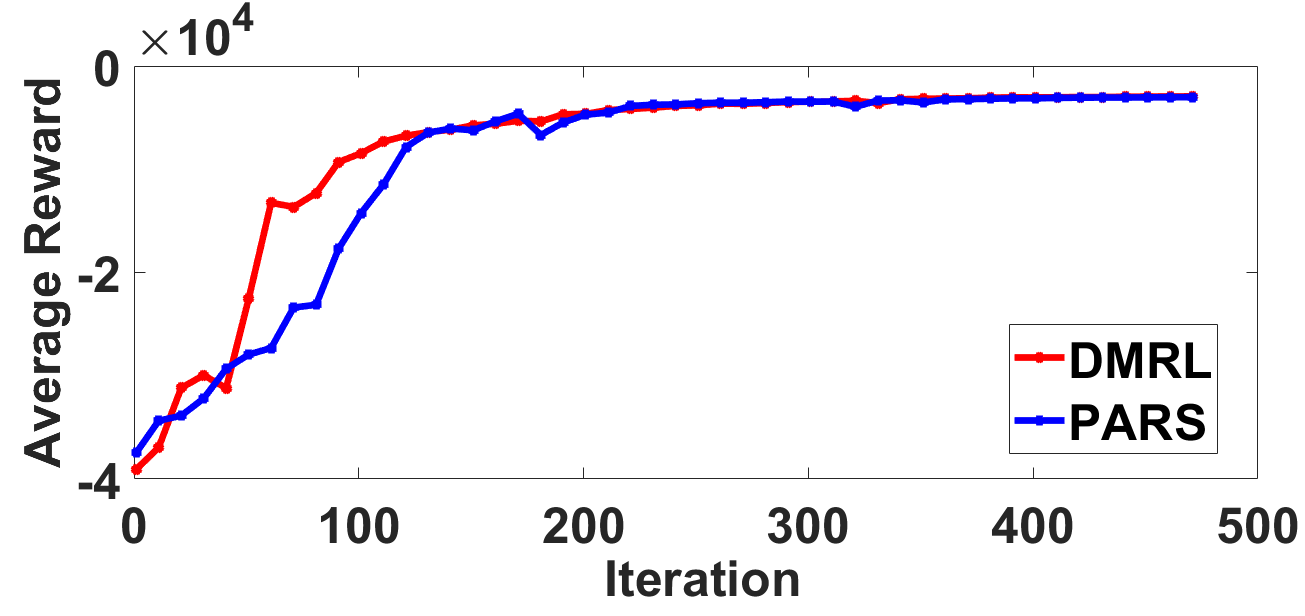}
\caption{Learning curves of DMRL and PARS.}
\label{learning_curve}
\end{figure}

The hyperparameters of the DMRL algorithm for the training is shown in Table \ref{tab_hyperpar}. Fig. \ref{learning_curve} shows the average rewards with respect to training iterations for both the DMRL and PARS algorithms. DMRL and PARS has very similar convergence speed at the training stage. Table \ref{tab_testtiming} shows the training time for both the DMRL and PARS methods. While the meta-training using BO method costs 2 more hours, the DMRL method can meet the day-ahead training requirement. The day-ahead meta-training produces one well-trained universal LSTM policy.

To comprehensively evaluate the proposed method, we perform adaptation and testing with 9 different power flow cases and 4 different set of dynamic models in this paper (a total of 36 different environments), as shown in Table \ref{tab_pfcases_testing} and Table \ref{tab_dyrcases}. Note that only one adaptation is usually needed for one system at one operation time instance in real-world application. We recognize that due to unpredicted and extreme situations, the load/generation pattern and the load dynamics of the specific target environments at the adaptation stage will not always be within the boundary considered in the training data set. Accordingly, we include some ``outliers" in Table \ref{tab_pfcases_testing} and Table \ref{tab_dyrcases}. We believe such a testing setup can help to comprehensively test the \textit {adaptation} and  \textit {generalization} capabilities of the proposed method.  

At the intra-hour adaptation stage, we assume the grid operators have a much better estimation of the load/generation pattern and of the load dynamics, but cannot accurately forecast the fault location nor its duration. To validate that the proposed adaptation \textbf{Algorithm 2} can work for a wide range of target environments that are different from the training environments, we defined a total of 36 different target environments for adaption $\lbrace\mathcal{E}_{adapt}\rbrace$, which are combinations of 9 different power flow conditions shown in Table \ref{tab_pfcases_testing} and 4 different sets of critical  dynamic load parameters shown in Table \ref{tab_dyrcases}.  Since we still cannot predict the fault location and duration accurately, for each target environment for adaptation $\mathcal{E}_{target,i}$, we sample the contingency set $\lbrace C_{sample}\rbrace$ and it has 102 different contingencies, which are combinations of 34 different fault buses and three different fault durations (0.06, 0.1 , and 0.18 second). Note that $\lbrace C_{sample}\rbrace$ is totally different from $\lbrace C_{train}\rbrace$. We generate different optimized latent vector for each different target environment in $\lbrace\mathcal{E}_{adapt}\rbrace$ separately. Table \ref{tab_testtiming} shows that the adaptation for one environment takes 14.1 minutes, demonstrating that our method can meet the intra-hour adaptation time requirement.

At the testing stage, as shown in Table \ref{tab_pfcases_testing}, we assume the testing environment still has some small (e.g., 2.1\% to 2.5\% \cite{li2020novel})  differences in terms of load, generation and net load when compared with the specific target environment we consider at the adaptation stage. Note that there is a one-to-one correspondence between adaptation and testing in terms of power flow condition, as shown in Table \ref{tab_pfcases_testing}. Therefore, we have a total of 36 specific target environments for testing and evaluation. For each target environment, we test the corresponding optimized control strategy on the contingency set $\lbrace C_{test}\rbrace$ that has 40 different contingencies with different fault buses and the fault duration being randomly sampled within the range between 0.05 second and 0.2 second. In summary, at the testing stage, we test the proposed DMRL method on a total of $36\times40=1440$ different test scenarios.

\begin{table}[t]
\caption{Hyperparameters for training 300-bus system}
\begin{center}
\begin{tabular}{|l|c|c|}
\hline
\textbf{Parameters}  & \textbf{300-Bus} \\

\hline
Policy Model & LSTM \\
\hline
Policy Network Size (Hidden Layers) & [64,64] \\
\hline
Number of Directions ($N$)  & 128 \\
\hline
Top Directions ($b$)  & 64 \\
\hline
Step Size ($\alpha$)  & 1\\
\hline
Std. Dev. of Exploration Noise ($\upsilon$)  & 2 \\
\hline
Decay Rate ($\varepsilon$) & 0.996 \\
\hline
BO iterations ($N_{BO}$) & 32 \\
\hline
Inner loop iteration ($N_{in}$) & 20 \\
\hline
Outer loop iteration ($N_{out}$) & 25 \\
\hline
Sampled environments ($K$) & 3 \\
\hline
Sampled contingencies ($Q$) & 12 \\
\hline
Sampled scenarios ($M$) & 36 \\
\hline
\end{tabular}
\label{tab_hyperpar}
\end{center}
\end{table}

\subsection{Performance evaluation}

Table {\ref{tab_testtiming}} shows the average computation time of the DMRL, PARS, and MPC methods for determining the control actions for each testing scenario. The event duration for each testing scenario is 8 s. As discussed in Section IV.A, all the three methods (DMRL, PARS, and MPC) obtain the observations from the grid simulation environment and provide control actions back to it every 0.1 s. Thus, there are a total of 80 control steps with 0.1 s time step. The “testing time” listed in the Table {\ref{tab_testtiming}} is the average “total computation time” for the whole 80 control steps for each testing scenario. That is, both DMRL and PARS take 0.009 s on average to compute control actions within the 0.1 s control interval. Therefore, both methods meet the “real-time” requirements for the load shedding actions for the FIDVR problem in real system. In contrast, the MPC method takes 0.79 s computational time to determine the actions per control action step, which does not meet the “real-time” requirements for the load shedding actions. The reason is that performing neural network forward pass to infer actions in the DMRL and PARS approaches is much more efficient than solving a complex optimization problem with the MPC method.

\begin{table}[t]
\caption{Comparison of computation time for PARS, DRML, and MPC}
\begin{center}
\begin{tabular}{|c|c|c|c|c|c|}
\hline
\textbf{ } & \textbf{Required} & \textbf{DMRL} & \textbf{PARS} & \textbf{MPC} \\
\hline
{\textbf{Training}} & $<$ 24 hours  & 11.6 hours  & 9.5 hours & -- \\ 
\hline
{\textbf{Adaptation}} & $<$ 60 mins  & 14.1 mins  & -- & --\\ 
\hline
{\textbf{Testing}} & $<$ 1 s  & 0.72 s  & 0.72 s & 63.31 s \\ 

\hline
\end{tabular}
\label{tab_testtiming}
\end{center}
\end{table}

\begin{figure}[t]
\centering
\includegraphics[width=\columnwidth]{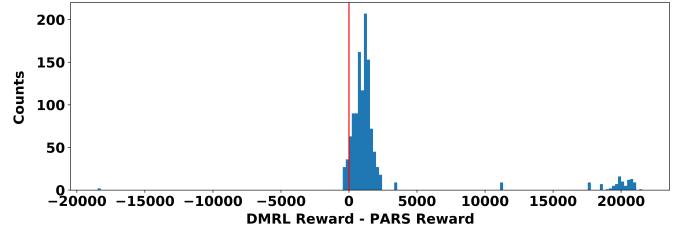}
\caption{Histogram of the reward differences between DMRL and PARS.}
\label{300_reward_diff}
\end{figure}

To show the advantage of our DMRL method, we calculated the reward differences (i.e., the reward of DMRL subtracts that of PARS ) for all 1440 different test scenarios. A positive value means the proposed DMRL method performs better for the corresponding Fig. \ref{300_reward_diff} shows the histogram of rewards differences for the 1440 different scenarios. Among 94.69\% of those scenarios, DMRL outperforms PARS since most of the reward differences are positive when compared with PARS. This confirms that with a learnt latent vector for each target environment, DMRL-based agent can better distinguish each environment and tailor the control strategy accordingly, thereby achieving overall better control effectiveness and performance. Furthermore,  there are 88 scenarios where DMRL successfully recovered the system voltage above the performance requirement shown in Fig. \ref{fig:volt_performance_curve} while PARS failed to do so. These scenarios are corresponding to a cluster of scenarios with reward difference of about +20000 in Fig. \ref{300_reward_diff}. Indeed, this result demonstrates the unwarranted generalization capability of DRL algorithms for some new or unseen grid environments while our DMRL algorithm can successfully fast adapt to unseen (even very different from those in the training set) grid operation environments.  

\begin{figure}[t]
\centerline{\includegraphics[scale=0.2]{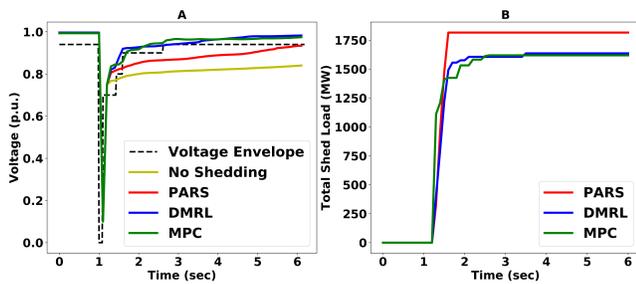}}
\caption{Test results for a FIDVR event triggered by a three-phase fault at bus 26 of the IEEE 300-bus system. (A) Voltage of bus 32. The dash line denotes the performance requirement for voltage recovery. (B) Total load shedding amount.}
\label{300-voltage-load}
\end{figure}

Fig. \ref{300-voltage-load} shows the comparison of DMRL, PARS and MPC performance for a test scenario with a three-phase fault at bus 26 lasting 0.15 second. The total rewards of PARS, DMRL and MPC methods are -25833, - 2715 and - 2742, respectively. Fig. {\ref{300-voltage-load}}(A) shows that the voltage with PARS (red curve) at bus 32 could not recover to meet the standard (dashed black curve), while with DMRL (blue curve) and MPC (green curve), voltages could be recovered to meet the performance standard. Further investigation indicated there were 11 buses that could not recover their voltage with the PARS method, while the DMRL and MPC methods successfully recovered all bus voltages above the required levels. Fig. {\ref{300-voltage-load}}(B) shows that the load shedding amount with the DMRL method is less than that of PARS, while being the same as the MPC method, suggesting that DMRL can achieve (near) optimal solutions with fast adaptation. 

%\subsection{adapt to novel reward functions-constraint or limit changes}

\section{Conclusions}\label{sec:conclusions}
In this paper, we developed a novel DMRL algorithm to enable DRL agents to learn a latent context from prior learning experience and leverage it to quickly adapt to new (i.e., unseen during training time) grid environments, which could be new power flow conditions, dynamic parameters or combinations of them. The key idea in our method is integrating meta strategy optimization (MSO), which is a latent-space based meta-learning algorithm, with one state-of-the-art parallel ARS (PARS) algorithm. In addition, the meta-training and adaption procedures fit well into the execution time frames of the existing grid operation practice. We demonstrate the performance of the DMRL algorithm on a variety of different scenarios with the IEEE 300-bus system and compare it with PARS and MPC methods. The DMRL algorithm can successfully adapt to new power flow conditions and system dynamic parameters in less than 15 minutes. The performance of the adapted DMRL-based control polices is notably better than that of the PARS algorithm, and comparable to that of the MPC solutions. Furthermore, our method can provide control decisions in real-time while MPC fails to do so.

Future research topics of interest include: 
\begin{enumerate}
     \item Application of the developed technique for other grid stability controls and in other areas such as distribution system control, microgrid and building management where adaptation of control strategies is also a key issue.  
    \item Transferring learnt control policies for one area to a different area of the power system or even a  different power system.
\end{enumerate}

% or


% \appendices

% % you can choose not to have a title for an appendix
% % if you want by leaving the argument blank
% \section{}

%\section*{Acknowledgment}


\ifCLASSOPTIONcaptionsoff
  \newpage
\fi

\bibliographystyle{IEEEtran}
\bibliography{references}


\comment{
\begin{IEEEbiography}{Michael Shell}
Biography text here.
\end{IEEEbiography}

\begin{IEEEbiographynophoto}{John Doe}
Biography text here.
\end{IEEEbiographynophoto}

%\newpage

\begin{IEEEbiographynophoto}{Jane Doe}
Biography text here.
\end{IEEEbiographynophoto}
}

%\vfill


\end{document}